\PassOptionsToPackage{table}{xcolor}
\documentclass[journal]{IEEEtran}
\usepackage{color}
\usepackage{multirow}
\usepackage{tmi}
\usepackage{cite}
\usepackage{amsmath,amssymb,amsfonts}
\usepackage{algorithm}
\usepackage{algorithmic}
\usepackage{graphicx}
\usepackage{textcomp}
\usepackage{verbatim}
\usepackage[colorlinks,linkcolor=red]{hyperref}
\usepackage{color}
\usepackage{makecell}
\usepackage{stfloats}
\usepackage{bbding}
\usepackage{framed}
\usepackage{latexsym}
\usepackage{url}
\usepackage[table]{xcolor}         
\usepackage{pifont}
\usepackage{arydshln}
\usepackage{bm}

\usepackage{booktabs}
\usepackage{multicol}
\usepackage{colortbl}
\usepackage{threeparttable}
\usepackage{wrapfig}

\usepackage[accsupp]{axessibility}  
\usepackage{xcolor}
\usepackage{orcidlink}
\newcommand{\methodName}{\textbf{S\&D-Messenger}}{}

\def\BibTeX{{\rm B\kern-.05em{\sc i\kern-.025em b}\kern-.08em
    T\kern-.1667em\lower.7ex\hbox{E}\kern-.125emX}}
\makeatletter
\DeclareRobustCommand\onedot{\futurelet\@let@token\@onedot}
\def\@onedot{\ifx\@let@token.\else.\null\fi\xspace}

\makeatother

\begin{document}
\title{S\&D Messenger: Exchanging Semantic and Domain Knowledge for Generic Semi-Supervised Medical Image Segmentation}
\author{Qixiang Zhang*, Haonan Wang*, and Xiaomeng Li, \IEEEmembership{Member, IEEE}
\thanks{* indicates equal contribution.}
\thanks{Q. Zhang, H. Wang, and X. Li are with the Department of Electronic and Computer Engineering, The Hong Kong University of Science and Technology, Hong Kong SAR, China (Corresponding author: Xiaomeng Li. e-mail: eexmli@ust.hk)}

\thanks{Copyright (c) 2022 IEEE. Personal use of this material is permitted. Permission from IEEE must be obtained for all other uses, including reprinting/republishing this material for advertising or promotional purposes, collecting new collected works for resale or redistribution to servers or lists, or reuse of any copyrighted component of this work in other works.}
}
\maketitle

\begin{abstract}
Semi-supervised medical image segmentation (SSMIS) has emerged as a promising solution to tackle the challenges of time-consuming manual labeling in the medical field. However, in practical scenarios, there are often domain variations within the datasets, leading to derivative scenarios like semi-supervised medical domain generalization (Semi-MDG) and unsupervised medical domain adaptation (UMDA). In this paper, we aim to develop a generic framework that masters all three tasks. We notice a critical shared challenge across three scenarios: \textit{the explicit semantic knowledge for segmentation performance and rich domain knowledge for generalizability exclusively exist in the labeled set and unlabeled set respectively}. Such discrepancy hinders existing methods from effectively comprehending both types of knowledge under semi-supervised settings. To tackle this challenge, we develop a \textbf{Semantic \& Domain Knowledge Messenger (S\&D Messenger)} which facilitates direct knowledge delivery between the labeled and unlabeled set, and thus allowing the model to comprehend both of them in each individual learning flow.
Equipped with our S\&D Messenger, a naive pseudo-labeling method can achieve huge improvement on \textbf{six benchmark datasets} for SSMIS (\textcolor{green}{+7.5\%}), UMDA (\textcolor{green}{+5.6\%}), and Semi-MDG tasks (\textcolor{green}{+1.14\%}), compared with state-of-the-art methods designed for specific tasks.
\end{abstract}

\begin{IEEEkeywords}
Medical image segmentation, Unsupervised Domain Adaptation, Domain Generalization, Semi-supervised Learning
\end{IEEEkeywords}

\section{Introduction}
\label{sec:introduction}

Semi-supervised Medical Image Segmentation (SSMIS) methods~\cite{bai2023bcp,gao2023correlation,wang2023dhc}, which leverage the principles of semi-supervised learning~\cite{MT,pseudo_label}, have been extensively researched as a means to train models using a limited amount of laborious and expensive labeled medical data, along with a larger set of readily available unlabeled data.
The success of SSMIS methods has paved the way for their application in more challenging scenarios. One such scenario is Unsupervised Medical Domain Adaptation (UMDA)~\cite{zhu2017cyclegan_uda,hoffman2018cycada_uda,tsai2018adaouput_uda}, where training data originates from two domains, and the target domain lacks accessible labels. 
Another scenario is Semi-supervised Medical Domain Generalization (Semi-MDG)\cite{liu2021dgnet,liu2021SDNet,yao2022epl,liu2022vmfnet}, where training data is obtained from multiple domains, with only a limited number having labeled data. It involves testing data from an unseen domain, creating a more challenging and realistic evaluation setting.

\begin{figure*}[t]
\centering
\includegraphics[width=0.95\textwidth]{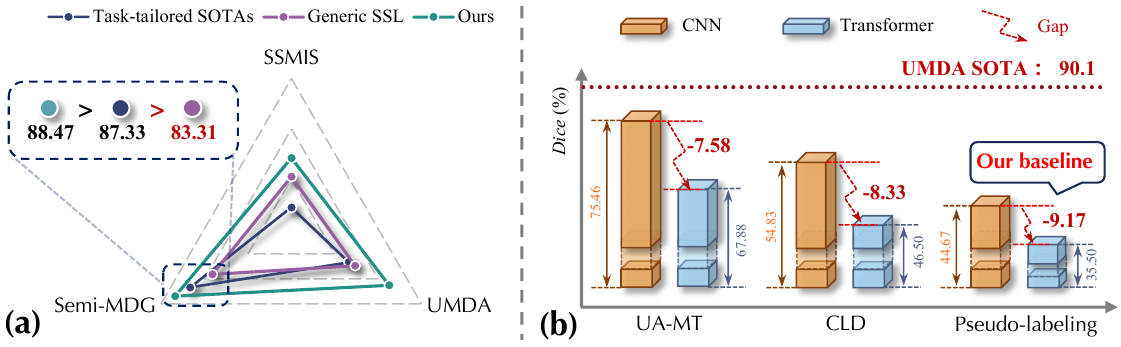}
\caption{\textbf{(a)}:  Compared with SOTA methods tailored for the specific scenario, GenericSSL~\cite{wang2024towards} achieved great improvement on SSMIS (\textcolor{green}{+11.85\%} Dice), however the results on UMDA (\textcolor{green}{+4.9\%} Dice) and Semi-MDG (\textcolor{red}{-2.15\%} Dice) can only match or even lag behind. \textbf{(b)}: Performance of typical SSMIS methods (UA-MT~\cite{yu2019uamt}, CLD~\cite{lin2022cld}, and Pseudo-labeling~\cite{lee2013pseudo}) on UMDA task (MMWHS dataset) with different backbones. These methods, with or without Transformer-based backbone, \textit{i.e.}, SegFormer~\cite{xie2021segformer}, encounter severe performance drops on the UMDA task, showing a large margin compared with the task-tailored SOTA.}
\label{intro:intro_bar}
\end{figure*}


Recently, Wang et al.~\cite{wang2024towards} highlighted that a strong semi-supervised framework can solve these three challenging tasks simultaneously. They coined the term ``generic semi-supervised medical image segmentation'' to describe this challenging yet meaningful setting.  
While their GenericSSL framework has achieved notable improvements in SSMIS tasks, with gains of approximately 12\% in Dice score, its advancements in tasks involving domain shift, \textit{e.g.}, UMDA and Semi-MDG, remain limited, compared to prior art methods that are specifically designed for these tasks. Specifically, the improvements in UMDA and Semi-MDG are only a modest increase of \textcolor{green}{1.0\%} or a decrease of \textcolor{red}{2.15\%} (refer to Fig.~\ref{intro:intro_bar}a).
Furthermore, when applied to UMDA and Semi-MDG tasks, other state-of-the-art (SOTA) SSMIS methods~\cite{yu2019uamt,lin2022cld} exhibited even inferior performance compared to GenericSSL; see results in Tab.~\ref{sota_UMDA},\ref{sota_SemiDG}.

One potential solution to mitigate the adversarial influence of domain shift is to adopt the transformer-based networks, \textit{e.g.}, SegFormer~\cite{xie2021segformer} with stronger generalizability~\cite{zhang2022delving}, rather than conventional CNNs to existing SSMIS framework~\cite{pseudo_label,yu2019uamt,lin2022cld}. However, based on our observation, when equipped with SegFormer, their performance somehow even suffers a more significant decrease; see results in Fig.~\ref{intro:intro_bar}b.

\begin{figure*}[t]
\centering
\includegraphics[width=\textwidth]{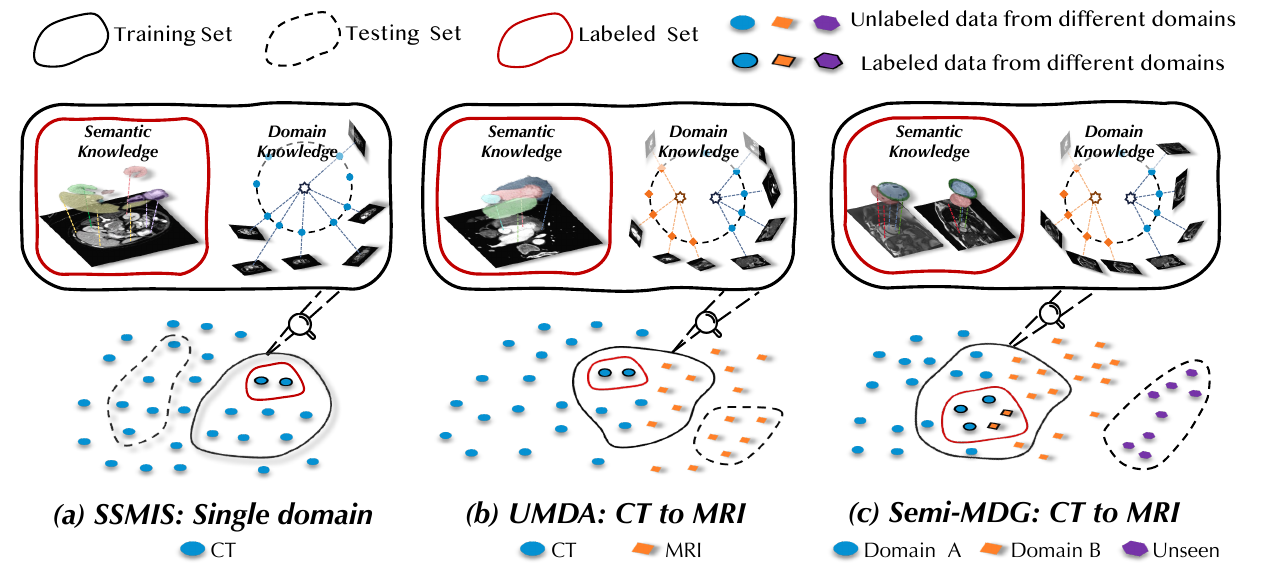}
\caption{The small labeled sets have explicit semantic knowledge derived from class-wise labels but lack comprehensive domain knowledge, \textit{i.e.}, feature variations across different domains, due to insufficient or entirely absent data from the target domain. Conversely, the large unlabeled sets exhibit the opposite characteristics, with abundant domain knowledge but lacking explicit semantic information.}
\label{intro:lab_unlab_knowledge}
\end{figure*}

The major reason for these above phenomenons is that previous methods (including GenericSSL~\cite{wang2024towards}) cannot simultaneously capture the \textit{semantic knowledge} for good segmentation performance and the \textit{domain knowledge} for good generalizability under the semi-supervised setting. 
As illustrated in Fig.~\ref{intro:lab_unlab_knowledge}, the segmentation annotations carrying explicit and precise \textit{semantic knowledge} only exist in the labeled set, while only the unlabeled set with a substantial volume of data from multiple domains could provide comprehensive and diverse \textit{domain knowledge}. However, recent methods~\cite{wang2024towards,chen2021cps,lin2022cld,bai2023bcp} neglect the potential relationship between the labeled set and the unlabeled set: the knowledge inside these two sets is complimentary. 
They separate the learning flow of the two sets, resulting in the model being exclusively exposed to a single type of knowledge during each individual learning flow. Consequently, the model tends to prioritize \textit{explicit} semantic knowledge, which can be easily acquired through \textit{precise} supervision, while largely neglecting the essential hidden domain knowledge required for comprehending domain patterns and cross-domain invariance. As a result, such discrepancy hampers the model's generalizability on tasks that involve multi-domain data, \textit{e.g.}, UMDA and Semi-MDG, leading to subpar performances. When equipped with SegFormer, such adversarial effects become even more pronounced, since transformer-based models are more likely to over-fit when labeled data is limited~\cite{dosovitskiy2020image}, compared with CNN-based models. 

\begin{figure}[t]
    \centering
\includegraphics[width=0.4\textwidth]{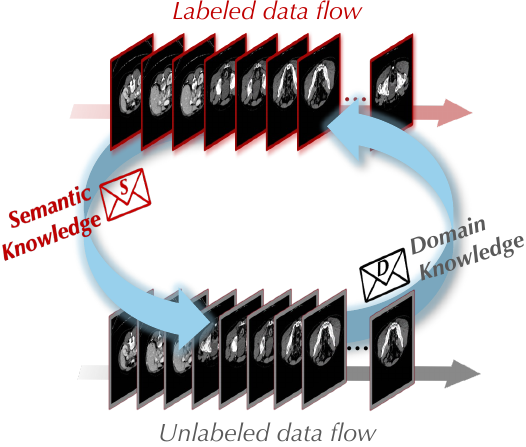}
    \caption{Insight of \methodName{}: enabling knowledge delivery across the learning flow of the labeled and unlabeled set. }\label{intro:semantic_messenger}
    \vspace{-0.5cm}
\end{figure}

To address this challenge, we propose \textbf{Semantic \& Domain Knowledge Messenger} (\methodName{}) break the barriers which separate the learning flows of the labeled and the unlabeled set by facilitating the delivery of complementary knowledge across these two flows; see Fig.~\ref{intro:semantic_messenger}. We employ a naive pseudo-labeling scheme~\cite{pseudo_label} as our baseline, where a single SegFormer~\cite{xie2021segformer} is trained with labeled data (with annotations) and unlabeled data (with current model predictions). Specifically, our \methodName{} consists of two knowledge delivery mechanisms. The Labeled-to-Unlabeled (L2U) delivery mixes a patch of the labeled foreground area with its multi-class annotation from the labeled flow to the unlabeled flow, which compensates for the absence of explicit semantic knowledge and precise supervision. The Unlabeled-to-Labeled (U2L) delivery replaces the original self-attention-based encoder in SegFormer with cross-attention-based \textit{\textbf{Messenger Transformer}} blocks, which considers the delivery destination (labeled channels) as \verb|queries| and the source (unlabeled channels) as \verb|keys| and \verb|values|. With these two knowledge delivery mechanisms, a naive pseudo-labeling framework could handle the core challenge of generic SSMIS and surpass the SOTA methods designed for specific scenarios; see Tab.~\ref{sota_IBSSL_20}-\ref{sota_scgm}.

The overall contributions can be summarized as follows:
\begin{itemize}
    \item We point out the major reason why previous methods perform badly in UMDA and Semi-MDG tasks: separate learning flows of labeled and unlabeled set that exclusively owns semantic knowledge and domain knowledge.
    \item We propose an efficient \methodName{} that facilitates knowledge exchange between that labeled and unlabeled set, which can also benefit the label-efficient training of any Transformer-based framework (\textit{e.g.}, SAM~\cite{liu2020saml}).
    \item Extensive experiments on \textbf{six representative datasets} of SSMIS, UMDA, and Semi-MDG tasks have validated the effectiveness of our method.
    Notably, our method achieves significant improvements in all three scenarios (7.5\% Dice on Synapse dataset, 5.6\% Dice on MMWHS dataset, and 1.14\% Dice on M\&Ms dataset).
\end{itemize}

\section{Related works}
\subsection{Semi-supervised Medical Image Segmentation}
Semi-supervised image segmentation aims to achieve comparable performance to fully-supervised methods using a small number of labeled samples and abundant unlabeled data.
Recently, self-training-based methods~\cite{chen2021cps,wang2022u2pl} have become the mainstream of this domain. Approaches with consistency regularization strategies~\cite{chen2021cps} achieved good performance by encouraging high similarity between the predictions of two perturbed networks for the same input image, which highly improved the generalization ability.
In the field of medical image analysis, the issue of limited data availability is particularly pronounced and poses a significant challenge.
Existing methods to combat the limited data include uncertainty suppression and consistency loss~\cite{luo2021urpc}, contrast learning sampling strategy by utilizing the most valuable knowledge from unlabeled data~\cite{wu2022cdcl}, rethinking Bayesian deep learning methods~\cite{wang2022gbdl}, exploring the pixel-level smoothness and inter-class separation~\cite{wu2022ssnet}, and improved Global Local CL~\cite{chaitanya2020glcl}.
Despite their notable successes, these methods face limitations when confronted with more challenging yet practical scenarios, such as UMDA and Semi-MDG. \\
\noindent \textbf{Transformer in Semi-supervised Segmentation.}
Recently, Vision Transformers (ViTs)~\cite{dosovitskiy2021vit,vitn_cai2023efficientvit} have remarkably reformed the field of image segmentation.
However, when it comes to semi-supervised settings, especially the medical image domain, it becomes challenging for such frameworks to achieve similar advances. The reason behind this lies in the weaker inductive bias of Transformer-based methods compared to CNN-based methods, as well as their heavy reliance on large amounts of training data~\cite{dosovitskiy2021vit,cai2022semivit}.
To address the over-fitting issue of ViTs, researchers from natural image and medical image domains have turned to a combination of ViTs and CNNs as a compromise~\cite{SemiCVT,DiverseCotraining,huang2023semi}. For example, SemiCVT~\cite{SemiCVT} introduces inter-model class-wise consistency to enhance the class-level statistics of CNNs and Transformers through cross-teaching. Additionally, other methods~\cite{DiverseCotraining,huang2023semi} employ ViTs as diverse models within the CPS framework~\cite{chen2021cps} alongside CNNs.
These hybrid methods have gained improved performance but have innate drawbacks: (1) CNNs' limited receptive field may influence Transformers' ability to capture long-range dependencies. (2) The hybrid methods result in complex network architectures, requiring intricate training strategies and parameter tuning.

Most recently, a \textit{concurrent} semi-supervised natural image segmentation work, AllSpark~\cite{allspark}, aims to enrich the output labeled features with unlabeled features to avoid the model being misled by the limited amount of labeled data. However, it operates only in the standard SSL \textbf{\textit{without considering domain shifts in the training}}. In contrast, our work tackles a more challenging scenario of GenericSSL, where training and test data exhibit domain shifts. Our U2L aims to address the domain shift by directly delivering domain knowledge to the encoder's hidden layers, thereby achieving better performance for GenericSSL (see Tab.~\ref{sota_IBSSL_20}-\ref{sota_scgm}).

It may bring some concerns that our L2U delivery module resembles BCP~\cite{bai2023bcp}. However, although our L2U and BCP both use classical copy-paste augmentation, the \textbf{motivation} of our L2U is to \textbf{\textit{provide}} more robust unlabeled features \textbf{\textit{for}} U2L to regularize labeled features, within a pure Transformer-based framework. In contrast, BCP applies copy-paste as an augmentation in the widely used mean teacher framework. Additionally, our method largely excels BCP (see Tab.~\ref{sota_IBSSL_20}-\ref{sota_scgm}).


\subsection{Unsupervised Medical Domain Adaptation}
Unsupervised Domain Adaptation~\cite{zhu2017cyclegan_uda,hoffman2018cycada_uda,tsai2018adaouput_uda} seeks to capture domain in-variance by training the model jointly using data originated from both source domain and target domain, while the target domain data lack any label. In the field of medical image segmentation, the two domains usually involve different modalities, \textit{e.g.}, MR, and CT images, where one of the modalities lacks any segmentation annotation.
UMDA has gained increasing attention in the medical imaging field because it offers an efficient way to compensate for limited medical image data. Consequently, numerous UMDA approaches have been developed for cross-domain medical image segmentation. These methods encompass various directions, \textit{e.g.}, semi-supervised learning~\cite{liu2022act_uda_new}, generative adversarial-based methods~\cite{dou2019pnp_uda,chen2020sifa_uda, zou2020dsfn_uda,han2021dsan_uda}, and contrastive learning~\cite{gu2022confuda_uda_new}. 

\subsection{Semi-supervised Medical Domain Generalization}
Semi-supervised domain generalization differs from unsupervised domain adaptation in that it does not require any data from the target domains, and the data in the source domains are only partially labeled. Overcoming this challenge necessitates the model to possess robust feature extraction capabilities and a high degree of generalizability.
Existing Semi-supervised domain generalization methods~\cite{liu2021dgnet,liu2021SDNet} use various carefully designed strategies to solve the domain shifts, \textit{e.g.}, meta-learning~\cite{liu2021dgnet}, and Fourier transformation~\cite{yao2022epl}.

As stated by Wang et al.~\cite{wang2024towards}, these three tasks sharing high similarity have been unified with a generic framework handling all of them. However, despite the advancement they achieved in SSMIS, the improvement on UMDA and Semi-MDG tasks is marginal compared with task-tailored SOTAs.

\begin{figure*}[!t]
\centering
\includegraphics[width=0.85\textwidth]{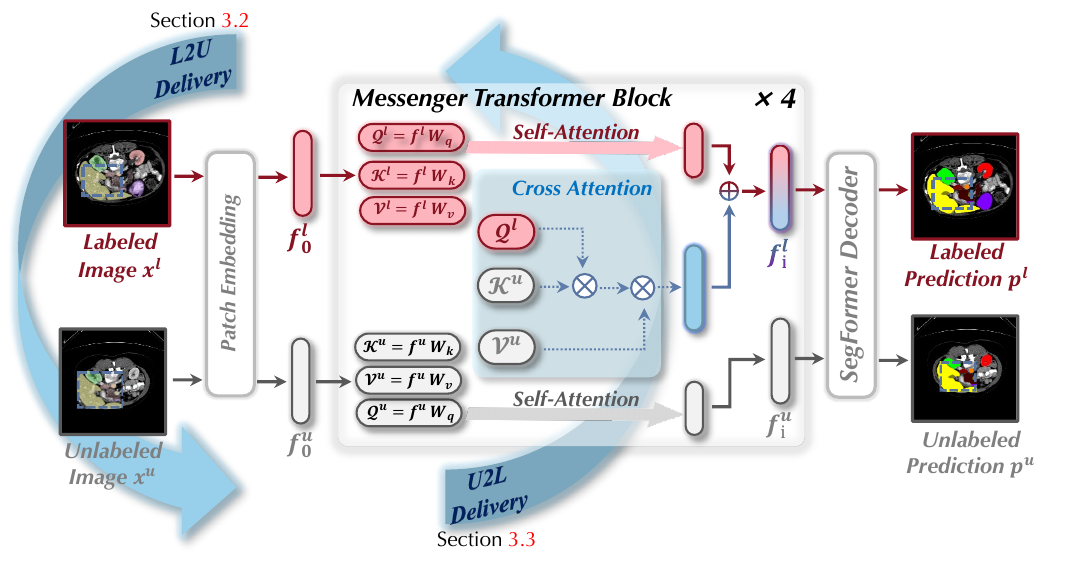}
\caption{Illustration of our proposed \methodName{} for SSMIS. Our \methodName{} inherits the architecture from pseudo-labeling with SegFormer as backbone~\cite{xie2021segformer}, while we replace the original Transformer Block with our \methodName{} which employs a novel feature-intervention attention module to use diverse unlabeled data to intervene in the learning flow of the labeled data.}
\label{method:framework}
\end{figure*}

\section{Methodology}
\subsection{Preliminary and Baseline}\label{sec:overview}
In our setting, we are provided with a small labeled set $\{(x_i^l,y_i)\}_{i=1}^{N_l}$ with $N_l$ labeled images and a relatively larger unlabeled set $\{x_i^u\}_{i=1}^{N_u}$ with $N_u$ unlabeled images ($N_l << N_u$).
$x_i \in \mathbb{R}^{3 \times H\times W}$ is the $i^{th}$ sample and $y_i \in \mathbb{R}^{L\times H\times W}$ is its corresponding ground-truth annotation with $L$ classes. 
Our method adopts SegFormer~\cite{xie2021segformer} to pseudo labeling framework~\cite{lee2013pseudo} as the foundation and baseline, where the unlabeled data is assigned with pseudo labels that are generated from current model predictions. The learning flow of the labeled data is supervised by ground truth annotation, while the generated pseudo labels $\hat{y}$ serve as supervision to the unlabeled learning flow. Therefore, the objective function of the baseline consists of a supervised loss $\mathcal{L}_{s}$ and an unsupervised loss $\mathcal{L}_{u}$:

\begin{scriptsize}
\begin{equation}
    \label{eq:loss}
    \mathcal{L} = \mathcal{L}_{s} + \mathcal{L}_{u}
    =\frac{1}{N_l}\sum_{i=0}^{N_l}\mathcal{L}_{CE}(f(x^{l}_i), y_i) + \frac{1}{N_u}\sum_{j=0}^{N_u} \mathcal{L}_{CE}(f(x^{u}_{j}), \hat{y}_{j})
\end{equation}
\end{scriptsize}

where $\mathcal{L}_{CE}$ denotes pixel-level cross-entropy loss, and $f$ is  segmentation model. 

In this paper, we take a close look at a shared essential challenge (see Fig.~\ref{intro:lab_unlab_knowledge}) existing in the generic SSMIS setting, and majorly focus on the architecture-level design.
We start with pseudo-labeling as our baseline and then propose two \textit{knowledge delivery} schemes, which together consist of our \methodName{}, as illustrated in Fig.~\ref{method:framework}.

\subsection{Labeled-to-Unlabeled (L2U) Knowledge Delivery}\label{sec:l2u}
The labeled set carries segmentation annotations that contain explicit semantic knowledge necessary for achieving good segmentation performance, which is notably absent in the unlabeled set due to the lack of annotation. Such absence severely hinders convergence to the unlabeled set, which further influences the extraction of the rich domain knowledge it contains. Therefore, we propose the Labeled-to-Unlabeled (L2U) Knowledge Delivery scheme, aiming to deliver the semantic knowledge from the learning flow of the labeled set to the unlabeled set. For each unlabeled sample $x_i^u$, we randomly select a labeled sample and its corresponding annotation $(x_j^l,y_j)$, and slice a patch $p$ that includes foreground areas of various classes, and then incorporate it into $x_i^u$, while the pixel-level annotation of the patch is also pasted in the pseudo-label $\hat{y}_{i}$ to serve as supervision:

\begin{footnotesize}
\begin{equation}
    \label{eq:loss_l2u_1}
    \begin{split}
        x_i^u[:, p_h:p_h+s, p_w:p_w+s] = x_j^l[:, p_h:p_h+s, p_w:p_w+s] \\
    \hat{y}_{i}[:, p_h:p_h+s, p_w:p_w+s] = y_j[:, p_h:p_h+s, p_w:p_w+s]
    \end{split}
\end{equation}
\end{footnotesize}

where $p_h, p_w$ denotes the horizontal and vertical coordinates of the top-left corner of $p$, $s$ denotes the patch size.
Consequently, we revise the loss function of the unlabeled set into the following:
\begin{equation}
    \label{eq:loss_l2u_2}
    \mathcal{L}_{u}
    =\frac{1}{N_u}\sum_{j=0}^{N_u} \mathcal{L}_{CE}(f({x^{u}_{j}}'), \hat{y}_{j}')
\end{equation}
where ${x^{u}_{j}}'$ and $\hat{y}_{j}'$ denotes the unlabeled sample and its corresponding pseudo-label, that are mixed with $p$.

In this way, we deliver the explicit semantic knowledge from the labeled set to the unlabeled set by offering the learning flow of the unlabeled set with precise supervision. 

\begin{table*}[!b]
\scriptsize
\caption{
Results on Synapse dataset with 20\% labeled data for \textbf{SSMIS} task. \textbf{Bold} and \underline{underline} denote the best and the second best results.
}
\label{sota_IBSSL_20}
\setlength\tabcolsep{3pt} \centering
\resizebox*{0.88\linewidth}{!}{
\begin{tabular}{l|cc|ccccccccccccc}
\toprule

\multirow{2}{*}{Methods}  & \multirow{2}{*}{Avg. Dice} & \multirow{2}{*}{Avg. ASD} & \multicolumn{13}{c}{Dice of Each Class}  \\ 
&                        &                          & Sp   & RK   & LK   & Ga   & Es   & Li   & St   & Ao   & IVC  & PSV  & PA   & RAG  & LAG  \\\midrule
V-Net (fully)      & $62.09_{\pm1.2}$	&$10.28_{\pm3.9}$	& 84.6	& 77.2	& 73.8	& 73.3	& 38.2	& 94.6	& 68.4	& 72.1	& 71.2	& 58.2	& 48.5	& 17.9	& 29.0 \\
SegFormer (fully)      & $72.29_{\pm1.7}$	&$5.33_{\pm1.6}$	& 94.6	& 87.2	& 83.4	& 74.2	& 83.5	& 94.5	& 85.4	& 73.6	& 74.2	& 68.7	& 58.5	& 34.3	& 27.6 \\
Sup Only      & $23.99_{\pm7.3}$	&$74.26_{\pm5.9}$	& 25.0	& 27.1	& 9.3	& 4.4	& 43.4	& 85.2	& 54.0	& 48.8	& 14.7	& 0.1	& 0.0	& 0.0	& 0.0 \\
\midrule

UA-MT~\cite{yu2019uamt}   & $20.26_{\pm2.2}$	&$71.67_{\pm7.4}$	& 48.2	& 31.7	& 22.2	& 0.0	& 0.0	& 81.2	& 29.1	& 23.3	& 27.5	& 0.0	& 0.0	& 0.0	& 0.0  \\
                 
URPC~\cite{luo2021urpc}      & $25.68_{\pm5.1}$	&$72.74_{\pm15.5}$	& 66.7	& 38.2	& 56.8	& 0.0	& 0.0	& 85.3	& 33.9	& 33.1	& 14.8	& 0.0	& 5.1	& 0.0	& 0.0  \\

CPS~\cite{chen2021cps}     & $33.55_{\pm3.7}$	&$41.21_{\pm9.1}$	& 62.8	& 55.2	& 45.4	& 35.9	& 0.0	&91.1	& 31.3	& 41.9	& 49.2	& 8.8	& 14.5	& 0.0	& 0.0  \\

SS-Net~\cite{wu2022ssnet}  & $35.08_{\pm2.8}$	&$50.81_{\pm6.5}$	& 62.7	& 67.9	& 60.9	& 34.3	& 0.0	& 89.9	& 20.9	& 61.7	& 44.8	& 0.0	& 8.7	& 4.2	& 0.0  \\

DST~\cite{chen2022dst}      & $34.47_{\pm1.6}$	&$37.69_{\pm2.9}$	& 57.7	& 57.2	& 46.4	& 43.7	& 0.0	& 89.0	& 33.9	& 43.3	& 46.9	& 9.0	& {21.0}	& 0.0	& 0.0  \\

DePL~\cite{wang2022depl}       & $36.27_{\pm0.9}$	&$36.02_{\pm0.8}$	& 62.8	& 61.0	& 48.2	& 54.8	& 0.0	& {90.2}	& {36.0}	& 42.5	& 48.2	& 10.7	& 17.0	& 0.0	& 0.0  \\ \midrule

Adsh~\cite{guo2022adsh}      & $35.29_{\pm0.5}$	&$39.61_{\pm4.6}$	& 55.1	& 59.6	& 45.8	& 52.2	& 0.0	& 89.4	& 32.8	& 47.6	& 53.0	& 8.9	& 14.4	& 0.0	& 0.0  \\ 

CReST~\cite{wei2021crest}      & $38.33_{\pm3.4}$	&$22.85_{\pm9.0}$	& 62.1	& 64.7	& 53.8	& 43.8	& {8.1}	& 85.9	& 27.2	& 54.4	& 47.7	& 14.4	& 13.0	& {18.7}	& 4.6  \\

SimiS~\cite{simis}      & $40.07_{\pm0.6}$	&$32.98_{\pm0.5}$	& 62.3	& {69.4}	& 50.7	& 61.4	& 0.0	& 87.0	& 33.0	& 59.0	& {57.2}	& {29.2}	& 11.8	& 0.0	& 0.0  \\

Basak \textit{et al.}~\cite{basak2022addressing}       &$33.24_{\pm0.6}$	&$43.78_{\pm2.5}$	& 57.4	& 53.8	& 48.5	& 46.9	& 0.0	& 87.8	& 28.7	& 42.3	& 45.4	& 6.3	& 15.0	& 0.0	& 0.0   \\
                                 
CLD~\cite{lin2022cld}      &$41.07_{\pm1.2}$	&$32.15_{\pm3.3}$	& 62.0	& 66.0	& {59.3}	& {61.5}	& {0.0}	& 89.0	& 31.7	& {62.8}	& 49.4	& 28.6	& 18.5	& {0.0}	& {5.0}  \\
 
DHC~\cite{wang2023dhc}   & $48.61_{\pm0.9}$	&$10.71_{\pm2.6}$	& 62.8	& 69.5	& 59.2	& \textbf{66.0}	& 13.2	& 85.2	& 36.9	& 67.9	& 61.5	& 37.0	& 30.9	& 31.4	& 10.6 \\ 

AllSpark~\cite{allspark}   & $60.68_{\pm0.6}$	&$2.37_{\pm0.3}$	& 86.3 & 79.6 & 77.8  & 60.4  & 60.7 &92.3 &63.7 &75.0 &69.9 &60.2 &57.7 &0.0 &5.2 \\

GenericSSL~\cite{wang2024towards}   & $60.88_{\pm0.7}$	&$2.52_{\pm0.4}$	& 85.2 & 66.9 & 67.0  & 52.7  & 62.9 &89.6 &52.1 &83.0 &74.9 &41.8 &43.4 &\textbf{44.8} &\underline{27.2} \\

\rowcolor{cyan!20} \textbf{Ours}   & {$\underline{64.46}_{\pm0.7}$}	&{$\underline{2.21}_{\pm0.4}$}	& \textbf{92.5} & \underline{83.4} & \underline{81.4}  & \underline{62.3}  & \textbf{68.8} &\textbf{95.7} &\underline{68.2} &\underline{81.1} &\textbf{76.5} &\textbf{64.3} &\textbf{63.1} &0.7 &0.0 \\

\rowcolor{cyan!20} \textbf{Ours}$^\dag$   & {$\textbf{68.38}_{\pm0.6}$}	&{$\textbf{2.16}_{\pm0.6}$}	& \underline{88.9} & \textbf{86.2} & \textbf{86.6}  & 45.1  & \underline{66.3} &\underline{94.4} &\textbf{73.6} &\textbf{83.0} &\underline{75.3} &\underline{60.5} &\underline{55.1} &\underline{34.1} &\textbf{39.8} \\

\bottomrule
\end{tabular}
}
\begin{threeparttable}
 \begin{tablenotes}
        \scriptsize
        \item[$^\dag$] Due to the highly imbalanced nature of this dataset, we follow~\cite{wang2024towards} to use the same class-imbalanced designs in our method.
\end{tablenotes}
\end{threeparttable}

\end{table*}
\subsection{Unlabeled-to-Labeled (U2L) Knowledge Delivery}
Although equipped with annotations that offer explicit semantic knowledge, the small labeled set lacks enough comprehensive domain knowledge which is crucial for good adaptation and generalization ability. The absence of domain knowledge makes it nearly impossible for the model to understand the underlying domain pattern and the cross-domain invariance from the learning flow of the labeled set, which may further lead to overwhelming overfitting and poor generalizability. Hence, we propose the Unlabeled-to-Labeled (U2L) Knowledge Delivery, aiming to deliver rich domain knowledge from the learning flow of the unlabeled set to the labeled set. Typically, different image feature channels encode representative information shared by various samples, making them an ideal source to extract common domain knowledge. Therefore, we propose to summarize the generic domain knowledge from channels of unlabeled features, and deliver it to the labeled flow through the channel-wise cross-attention mechanism~\cite{wang2022uctransnet}.
Within this mechanism, we utilize the unlabeled channels to regularize the labeled feature. Specifically, we calculate the similarity between the labeled channels (\verb|queries|) and the unlabeled channels (\verb|keys|, \verb|values|), and those channels with the highest similarities play a more significant role in the regularization. The underlying reasoning behind this is to leverage the unlabeled features sharing similar semantic information with the labeled feature to regularize its individual deviation which hinders the capture of domain pattern. Such delivery is embedded into every Transformer block inside the encoder, as illustrated in Fig.~\ref{method:framework}.

Given the hidden features of labeled data and unlabeled data at any arbitrary stage of the encoder, we split them and obtain the \verb|queries| $Q$ from the labeled feature $f^l$, the \verb|keys| $K$ and \verb|values| $V$ from the unlabeled feature $f^u$ as $Q,K,V=f^lW_Q, f^uW_K, f^uW_V$, where $W_{Q/K/V} \in \mathbb{R}^{C \times 2C}$ are projection matrix for \verb|queries|, \verb|keys|, \verb|values| respectively, $C$ is the channel number.
Then, we could obtain the regularization feature for ${f}^{l}$ with channel-wise cross attention:

\vspace{-0.4cm}
\begin{small}
\begin{equation}
\hat{f}^{l} = Attention(Q^\top,K^\top,V^\top)  = softmax[\psi(Q^\top K)] V^\top 
\label{eq:u2l_1}
\end{equation}
\end{small}

where $\psi(\cdot)$ denote the instance normalization~\cite{IN}. Then, the regularization feature $\hat{f}^{l}$ from the unlabeled set are then added to the original labeled feature to serve as the final output for the labeled flow:

\begin{small}
\begin{equation}
\widetilde{f}^{l} = [\alpha \times \hat{f}^{l} + (1-\alpha){f}^{l}] W_O
\label{eq:u2l_2}
\end{equation}
\end{small}

where $\alpha$ is a coefficient, $W_O \in \mathbb{R}^{2C \times C}$ is the output projection layer. Note that, to refine the original labeled feature ${f}^{l}$, it first goes through a channel-wise self-attention module before being regularized by $\hat{f}^{l}$. The self-attention formulation closely resembles Eq.~\ref{eq:u2l_1}, but with a minor distinction: $K$ and $V$ are derived from $K,V=f^lW_K, f^lW_V$. For the unlabeled flow, we also employ a channel-wise self-attention module to obtain the refined unlabeled feature $\widetilde{f}^{u}$. During inference, we simplify the cross-attention to self-attention, where the test data are doubled and go through both flows.

\begin{figure*}[t]
\centering
\includegraphics[width=0.92\textwidth]{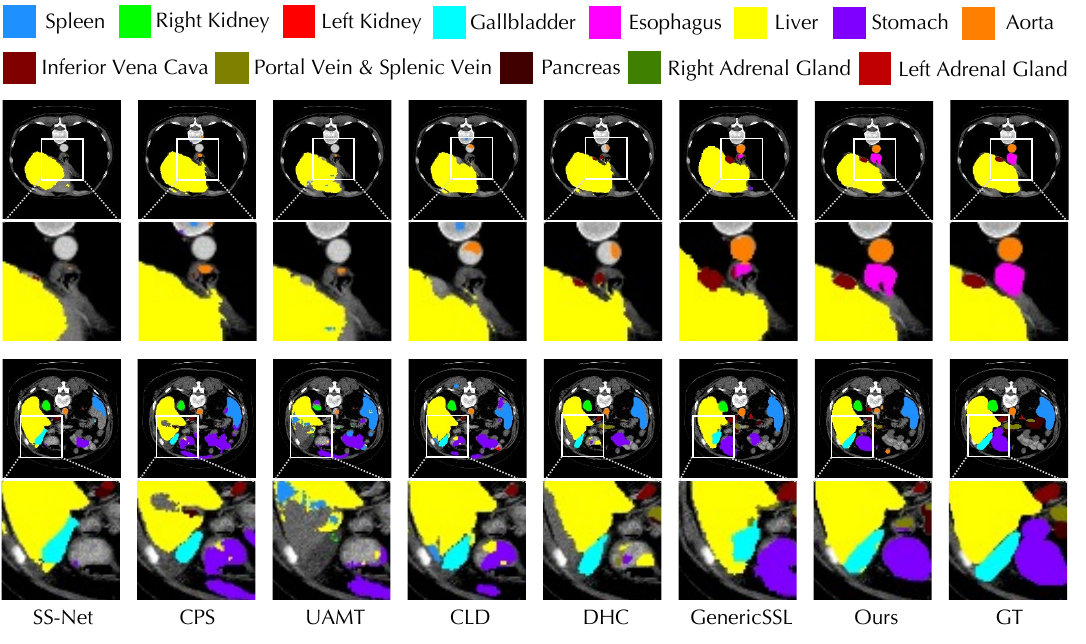}
\caption{Qualitative comparison between \methodName{} and the SOTA methods on 20\% labeled Synapse dataset.}
\label{synapse}
\vspace{-0.5cm}
\end{figure*}

\section{Experiment}
\subsection{Datasets and Implementation Details}
We evaluate our method on six datasets under the semi-supervised setting, \textit{i.e.}, the Synapse dataset~\cite{synapse}, LASeg dataset~\cite{xiong2021laseg}, and AMOS dataset~\cite{ji2022amos} for SSMIS, MMWHS dataset~\cite{zhuang2016mmwhs} for UMDA, M\&Ms dataset and SCGM dataset~\cite{prados2017spinal} for Semi-MDG. We evaluate the prediction of the network with two metrics, including Dice and the average surface distance (ASD). For LASeg dataset, we employ additional metrics, Jaccard and HD95, following previous works~\cite{wang2024towards}.

We implement the proposed framework with PyTorch 2.1.0, using one NVIDIA H800 GPU card. We employ the Stochastic Gradient Descent (SGD) optimizer and utilize polynomial scheduling to adapt the learning rate. The learning rate is adjusted using the formula  $lr = lr_{init} \cdot (1 - \frac{i} {I} )^{0.9} $, where $lr_{init}$ denotes the initial learning rate, $i$ is the current iteration, and $I$ is the maximum number of iterations. The batch size is set to 16. We consider a pseudo-labeling scheme~\cite{pseudo_label} with SegFormer-B5 as our baseline and backbone. The optimal values for $s$ in Eq.~\ref{eq:loss_l2u_1} and $\alpha$ in Eq.~\ref{eq:u2l_2} are examined in Tab.~\ref{ablation:hyparameter} through ablation analysis.

\subsection{Comparison with State-of-the-art Methods}

\begin{table}[h]
\caption{Results on LASeg dataset for \textbf{SSMIS} task.}
\label{sota_SSL}
\centering
\setlength\tabcolsep{7pt}
\resizebox*{0.44\textwidth}{!}{
\begin{tabular}{lcccc}
\toprule
\rowcolor{red!10}\multicolumn{5}{c}{5\% labeled data (labeled:unlabeled=4:76)} \\ \toprule
\multirow{2}{*}{Method} & \multicolumn{4}{c}{Metrics}  \\
    & Dice        & Jaccard        & 95HD & ASD \\ \midrule
SegFormer (fully) &   91.47  &  84.36   & 5.48 & 1.51  \\
Sup only (5\%) &   66.48  &  49.79   & 20.35 & 6.57 \\ \midrule
UA-MT~\cite{yu2019uamt} &   82.26  &  70.98   & 13.71     & 3.82   \\
SASSNet~\cite{li2020sassnet} &   81.60  &  69.63   & 16.16     & 3.58   \\
DTC~\cite{luo2021dtc} &   81.25  &  69.33   & 14.90     & 3.99   \\
URPC~\cite{luo2021urpc} &   82.48  &  71.35   & 14.65     & 3.65   \\
MC-Net~\cite{wu2021mcnet} &   83.59  &  72.36   & 14.07     & 2.70   \\
SS-Net~\cite{wu2022ssnet} &   86.33  &  76.15   & 9.97     & 2.31   \\
BCP~\cite{bai2023bcp} &   88.02  & 78.72    & 7.90    & 2.15  \\
AllSpark~\cite{allspark} & 87.99   & 78.83    &  7.44    & 2.10   \\
GenericSSL~\cite{wang2024towards} & \underline{89.93}   & \underline{81.82}    &  \underline{5.25}    & \underline{1.86}   \\

\rowcolor{cyan!20}\textbf{Ours} & \textbf{90.21}   & \textbf{82.65}    &  \textbf{4.74} & \textbf{1.63}   \\
\bottomrule
\toprule
\rowcolor{red!10}\multicolumn{5}{c}{10\% labeled data (labeled:unlabeled=8:72)} \\ \toprule

SegFormer (fully) &   91.47  &  84.36   & 5.48 & 1.51  \\
Sup only (10\%) &   84.81  &  73.62   & 11.63     & 3.14 \\ \midrule
UA-MT~\cite{yu2019uamt} &   87.79  &  78.39   & 8.68     & 2.12   \\
SASSNet~\cite{li2020sassnet} &   87.54  &  78.05   & 9.84     & 2.59   \\
DTC~\cite{luo2021dtc} &   87.51  &  78.17   & 8.23     & 2.36   \\
URPC~\cite{luo2021urpc} &   86.92  &  77.03   & 11.13     & 2.28   \\
MC-Net~\cite{wu2021mcnet} &   87.62  &  78.25   & 10.03     & 1.82   \\
SS-Net~\cite{wu2022ssnet} &   88.55  &  79.62   & 7.49     & 1.90   \\
BCP~\cite{bai2023bcp} &   89.62  &  81.31   & 6.81    & 1.76  \\

AllSpark~\cite{allspark} &88.74  &80.54 & 7.06 & 1.82   \\

GenericSSL~\cite{wang2024towards} &\underline{90.31}  &\underline{82.40} &\underline{5.55} &\underline{1.64}   \\

\rowcolor{cyan!20}\textbf{Ours} & \textbf{91.46}  &\textbf{83.55} &\textbf{5.02} &\textbf{1.33}   \\
\bottomrule
\end{tabular}
}
\begin{threeparttable}
 \begin{tablenotes}
        \scriptsize
        \item[$\star$] SOTA methods tailored for UMDA and Semi-MDG, respectively.\\
        \item[``-''] refers to not reported in the original papers.
\end{tablenotes}
\end{threeparttable}
\vspace{-0.4cm}
\end{table}



\begin{table}[!ht]
\caption{RESULTS ON AMOS DATASET FOR SSMIS TASK.}
    \label{amos}
    \setlength\tabcolsep{3pt}
    \centering
    \resizebox*{0.48\textwidth}{!}{
    \begin{tabular}{lccccc}
    \toprule
\rowcolor{red!10}\multicolumn{6}{c}{2\% labeled data (labeled:unlabeled=4:212)} \\ \toprule
Metrics & SegFormer (fully) & CPS~\cite{chen2021cps} & DHC~\cite{wang2023dhc} & AllSpark~\cite{allspark} & \cellcolor{cyan!20}{\textbf{Ours}}  \\ \midrule

Avg. Dice & $76.50_{\pm2.32}$ & $31.78_{\pm5.44}$ & $38.28_{\pm1.93}$ & $40.20_{\pm2.29}$ &  \cellcolor{cyan!20}{\textbf{$48.99_{\pm1.78}$}} \\

Avg. ASD & $2.01_{\pm1.47}$ & $39.23_{\pm7.24}$ & $20.34_{\pm4.22}$ & $14.77_{\pm2.88}$  &  \cellcolor{cyan!20}{\textbf{$10.70_{\pm3.90}$}} \\ 
\bottomrule 
\bottomrule
\rowcolor{red!10}\multicolumn{6}{c}{5\% labeled data (labeled:unlabeled=11:205)} \\ \toprule
Metrics & SegFormer (fully) & CPS~\cite{chen2021cps} & DHC~\cite{wang2023dhc} & AllSpark~\cite{allspark} & \cellcolor{cyan!20}{\textbf{Ours}}  \\ \midrule

Avg. Dice & $82.39_{\pm3.64}$ & $41.08_{\pm3.09}$ & $49.53_{\pm2.22}$ & $53.77_{\pm1.88}$ &  \cellcolor{cyan!20}{\textbf{$57.83_{\pm1.64}$}} \\

Avg. ASD & $1.19_{\pm0.67}$ & $20.37_{\pm2.97}$ & $13.89_{\pm3.64}$ & $10.96_{\pm2.28}$  &  \cellcolor{cyan!20}{\textbf{$7.85_{\pm2.74}$}} \\ 
\bottomrule 
    \end{tabular}}
    \vspace{-0.7cm}
\end{table}

\subsubsection{Results on SSMIS.}
We evaluate our proposed \methodName{} on three SSMIS benchmarks: Synapse, LASeg, and AMOS datasets.
We compare our method with several state-of-the-art SSMIS  methods~\cite{yu2019uamt,luo2021urpc,wu2022ssnet}. 
As shown in Tab.~\ref{sota_IBSSL_20}, Tab.~\ref{sota_SSL} and Tab.~\ref{amos}, our method outperforms all competing approaches on the Synapse, LASeg, and AMOS datasets. 
Notably, on the Synapse dataset, our method achieves SOTA segmentation performance on most types of organs, which shows the promising ability to solve traditional SSMIS tasks. Furthermore, by incorporating the class-imbalanced designs from~\cite{wang2024towards}, our method successfully segments the minor classes RAG and LAG, resulting in an improved overall Dice score of 68.38\% under the 20\% labeled setting. This achievement demonstrates a significant improvement (7.5\% in Dice) compared to previous methods.

Notably, we also present qualitative results in Fig.~\ref{synapse}, demonstrating that our method delivers more accurate and smooth segmentation predictions compared to other methods.

We further evaluate our method on the LASeg dataset, and the experimental results are shown in Tab.~\ref{sota_SSL}. Similarly, our method also achieves SOTA performance on the LASeg dataset. Specifically, as shown in Tab.~\ref{sota_SSL}, on the LASeg dataset with 10\% labeled data, our method outperforms the second-best method, \textit{i.e.}, GenericSSL~\cite{wang2024towards} with 1.15\% in Dice, 1.15 in Jaccard, and 0.53\% in 95HD.

\begin{table}[t]
\caption{Results on two settings, \textit{i.e.}, MR to CT and CT to MR, of MMWHS dataset for \textbf{UMDA} task.}
\label{sota_UMDA}
\setlength\tabcolsep{3pt}
\centering
\resizebox*{0.45\textwidth}{!}{
\begin{tabular}{lcccccc}
\toprule
\rowcolor{red!10}\multicolumn{7}{c}{MR to CT} \\ \toprule
\multirow{2}{*}{Method}    & \multicolumn{5}{c}{Dice} & \multicolumn{1}{c}{ASD}\\
& AA        & LAC        & LVC & MYO & Average  & Average \\ \midrule
SegFormer (Fully) &   92.7  &  91.1   & 91.9  & 87.8   &  90.9  &2.2 \\ \midrule
PnP-AdaNet~\cite{dou2019pnp_uda} &  74.0     & 68.9    & 61.9    & 50.8  &63.9    & 12.8  \\
AdaOutput~\cite{tsai2018adaouput_uda}    &     65.2      &      76.6      &  54.4   &  43.6   & 59.9       &9.6   \\
CycleGAN~\cite{zhu2017cyclegan_uda}    &     73.8      &      75.7      &  52.3   &  28.7   &  57.6       &10.8   \\
CyCADA~\cite{hoffman2018cycada_uda}    &     72.9      &      77.0      &  62.4   &  45.3   &   64.4       &9.4  \\
SIFA~\cite{chen2020sifa_uda}    &     81.3      &      79.5      &  73.8   &  61.6   &    74.1       &7.0\\
DSFN~\cite{zou2020dsfn_uda}    &     84.7      &      76.9      &  79.1   &  62.4   &    75.8       &- \\
DSAN~\cite{han2021dsan_uda}    &     79.9      &      84.8      &  82.8   &  66.5   &    78.5       &5.9     \\
LMISA-3D~\cite{jafari2022lmisa_uda}    &84.5 &82.8 &88.6 &70.1 &81.5 &2.3   \\
AllSpark~\cite{allspark}  &{87.0} &88.5 &86.4 &\underline{88.7} &87.6 &{2.0} \\
GenericSSL~\cite{wang2024towards}  &\textbf{93.2} &\underline{89.5} &\textbf{91.7} &{86.2} &\underline{90.1} &\underline{1.7} \\
\rowcolor{cyan!20} \textbf{Ours} &\underline{88.9} &\textbf{94.1} &\underline{88.5} &\textbf{94.2} &\textbf{91.4} &\textbf{1.4} \\
\bottomrule

\bottomrule
\rowcolor{red!10}\multicolumn{7}{c}{CT to MR} \\ \toprule
SegFormer (Fully) &   92.7  &  91.1   & 91.9  & 87.8   &  90.9  &2.2 \\ \midrule
PnP-AdaNet~\cite{dou2019pnp_uda} &  43.7     & 68.9    & 61.9    & 50.8  &63.9    & 8.9  \\
AdaOutput~\cite{tsai2018adaouput_uda}    &     60.8      &      39.8      &  71.5   &  35.5   & 51.9       &5.7   \\
CycleGAN~\cite{zhu2017cyclegan_uda}    &     64.3      &      30.7      &  65.0   &  43.0   &  50.7       &6.6   \\
CyCADA~\cite{hoffman2018cycada_uda}    &     60.5      &      44.0      &  77.6   &  47.9   &   57.5       &7.9  \\
SIFA~\cite{chen2020sifa_uda}    &     65.3      &      62.3      &  78.9   &  47.3   &    63.4       &5.7\\
DSAN~\cite{han2021dsan_uda}    &     {71.3} &66.2 &76.2 &52.1 &66.5 &5.4     \\
LMISA-3D~\cite{jafari2022lmisa_uda}    &60.7 &72.4 &\underline{86.2} &64.1 &70.8 &\underline{3.6}   \\
AllSpark~\cite{allspark}  &\underline{72.7} &73.7 &85.2 &63.8 &\underline{73.9} & 4.2 \\
GenericSSL~\cite{wang2024towards}  &62.8 &\underline{87.4} &61.3 &74.1 &{71.4} &7.9 \\
\rowcolor{cyan!20} \textbf{Ours} &\textbf{74.9} &\textbf{75.4} &\textbf{91.1} &\textbf{66.5} &\textbf{77.0} &\textbf{3.4} \\
\bottomrule
\end{tabular}}
\begin{threeparttable}
 \begin{tablenotes}
        \scriptsize
        \item[$\star$] SOTA methods on semi-supervised medical image segmentation.
        \item[``-''] refers to not reported in the original papers.
\end{tablenotes}
\end{threeparttable}
\vspace{-0.4cm}
\end{table}

\subsubsection{Results on UMDA.}

Tab.~\ref{sota_UMDA} shows the results of our method for UMDA task on the MMWHS dataset.
Existing UMDA methods mostly focused on using image-level (CycleGAN~\cite{zhu2017cyclegan_uda}), feature-level (DSAN~\cite{han2021dsan_uda}) or both (CyCADA~\cite{hoffman2018cycada_uda}, SIFA~\cite{chen2020sifa_uda}) alignments to mitigate the adversarial influence caused by the severe domain shifts. However, compared with the fully-supervised upper-bound, most of the other methods still produce subpar performance, especially on the ascending aorta (AA) and the myocardium of the left ventricle (MYO). Our method, on the contrary, achieves comparable performance with the upper-bound with only 0.5\% Dice margin on MR to CT task. Moreover, on difficult classes, \textit{e.g.}, AA, our method can still produce promising outcomes, which even surpass the upper bound, showing its extraordinary performance. 

\begin{table*}[t]
\caption{Results on 2\% and 5\% labeled data settings of M\&Ms dataset for \textbf{Semi-MDG} task. \textbf{Bold} and \underline{underline} denotes the best and the second-best results.}
\label{sota_SemiDG}
\setlength\tabcolsep{2pt}
\centering
\resizebox*{0.9\textwidth}{!}{
\begin{tabular}{c|cccc|c||cccc|c}
\toprule
\multirow{2}{*}{Method}         & \multicolumn{5}{c||}{2\% Labeled data}  & \multicolumn{5}{c}{5\% Labeled data}\\
& Domain A        & Domain B        & Domain C & Domain D & \textbf{Average} & Domain A        & Domain B        & Domain C & Domain D & \textbf{Average} \\ \midrule
nnUNet~\cite{isensee2021nnunet} &   52.87  &  64.63   &  72.97   &    73.27  &  65.94  &   65.30  &  79.73   & 78.06        & 81.25   &  76.09     \\
SDNet+Aug~\cite{liu2021SDNet}     &54.48     &      67.81      & 76.46   &  74.35   &    68.28  &71.21    &77.31  &81.40     &79.95     &77.47                \\
LDDG~\cite{li2020lddg}    &     59.47      &      56.16      &  68.21   &  68.56   &    63.16    &     66.22      &      69.49      &  73.40   &  75.66   &    71.29        \\
SAML~\cite{liu2020saml}    &     56.31      &      56.32      &  75.70   &  69.94   &    64.57   &     67.11      &      76.35      &  77.43   &  78.64   &    74.88         \\
BCP~\cite{bai2023bcp}$^\star$    &     71.57      &      76.20      &  76.87   &  77.94   &  75.65    &     73.66      &      79.04      &  77.01   &  78.49   & 77.05             \\
DGNet~\cite{liu2021dgnet}    &     66.01      &      72.72      &  77.54   &  75.14   &    72.85   &     72.40      &      80.30      &  82.51   &  83.77   &    79.75          \\
vMFNet~\cite{liu2022vmfnet}    &     73.13      &      77.01      &  81.57   &  82.02   &    78.43  &     77.06      &      82.29      &  84.01   &  85.13   &    82.12           \\
GenericSSL~\cite{wang2024towards}    &79.62 &82.26 &80.03 &83.31 &81.31 &81.71 &{85.44} &82.18 &83.90&83.31 
\\
AllSpark~\cite{allspark}    &83.77 &84.39 &85.29 &86.44 & 84.97 & 82.05 &84.39 &84.76 &84.27&86.36 
\\
Meta~\cite{liu2021semi} & 66.01 &72.72 & 77.54 & 75.14 &72.85 &72.40 & 80.30 & 82.51 &83.77 & 79.75 \\
StyleMatch~\cite{zhou2023semi} & 74.51 &77.69 & 80.01 & 84.19 & 79.10 &81.21 &  82.04 &  83.65 &83.77 & 82.67 \\
EPL~\cite{yao2022enhancing}    &82.35 &{82.84} &{86.31} &{86.58} &{84.52} &{83.30} &85.04 &{87.14} &{87.38} &{85.72} \\
$IS^2Net$~\cite{xie2023is2net}    &\underline{84.72} &\underline{84.97} &\underline{87.96} &\underline{89.01} &\underline{ 86.67} &\underline{85.11} &\underline{87.21} &\underline{87.49} &\underline{89.52} &\underline{ 87.33} \\
\rowcolor{cyan!20} \textbf{Ours} &\textbf{86.50} &\textbf{87.58} &\textbf{88.30} &\textbf{89.76} &\textbf{88.04} &\textbf{87.04} &\textbf{87.32} &\textbf{89.67} &\textbf{89.85} &\textbf{88.47}\\
\bottomrule
\end{tabular}}
\vspace{-0.5cm}
\begin{threeparttable}
 \begin{tablenotes}
        \scriptsize
        \item[$\star$] SOTA method on semi-supervised medical image segmentation.
\end{tablenotes}
\end{threeparttable}
\end{table*}

\begin{table}[!ht]
\caption{Results on SCGM dataset for \textbf{Semi-MDG} task. \textbf{Bold} and \underline{underline} denotes the best and the second best results.}
\label{sota_scgm}
\setlength\tabcolsep{3pt}
\centering
\resizebox*{0.48\textwidth}{!}{
\begin{tabular}{c|cccc|c}
\toprule
\rowcolor{red!10}Method                      & Domain A       & Domain B       & Domain C       & Domain D       & Average        \\
\midrule
FixMatch~\cite{sohn2020fixmatch}                      &  80.95         &  82.10          &  70.71          &  87.95          & 80.43          \\
CPS~\cite{chen2021cps}                      &  84.28         &   85.62          &   76.57          &   88.66          & 83.78          \\
nnUNet~\cite{isensee2021nnunet}                      & 59.07          & 69.94          & 60.25          & 70.13          & 64.85          \\
LDDG~\cite{li2020lddg}                        & 77.71          & 44.08          & 48.04          & 83.42          & 63.31          \\
SDNet+Aug~\cite{liu2021SDNet}                   & 83.07          & 80.01          & 58.57          & 85.27          & 76.73          \\
SAML~\cite{liu2020saml}                      & 78.71          & 75.58          & 54.36          & 85.36          & 73.50          \\
Meta~\cite{liu2021semi}                        & 87.45          & 81.05          & 61.85          & 87.96          & 79.58          \\
EPL~\cite{yao2022enhancing}                         & 87.13          & 87.31          & 78.75          & 91.73          & 86.23          \\
StyleMatch~\cite{zhou2023semi}                  & 82.59          & 83.26          & 72.14          & 88.01          & 81.50          \\
$IS^2Net$~\cite{xie2023is2net}                      & \underline{89.03}          & \underline{87.95}          & \underline{80.06}          & \textbf{92.38}          & 87.36          \\
\rowcolor{cyan!20} \textbf{Ours} & \textbf{89.77} & \textbf{88.40} & \textbf{83.22} & \underline{92.24} & \textbf{88.41} \scriptsize \\ \bottomrule
\end{tabular}}
\vspace{-0.5cm}
\end{table}

\subsubsection{Results on Semi-MDG.}
Tab.~\ref{sota_SemiDG} and Tab.~\ref{sota_scgm} respectively show the comparison results between our method and Semi-MDG methods, \textit{e.g.}, StyleMatch~\cite{zhou2023semi} and EPL~\cite{yao2022enhancing}, on M\&Ms dataset and SCGM dataset. 
These Semi-MDG methods, with various techniques tailored to deal with the domain shift, \textit{e.g.}, Fourier transformation~\cite{yao2022enhancing}, achieve improvements in the overall generalization performance. However, when in the face of large domain gaps, \textit{e.g.}, to domain A on M\&Ms dataset, the effectiveness becomes slight and minor. 
In contrast, our method proposes a more effective benchmark that leverages the representative transformer to learn more general representations of various domains. Thus, even without any special design to deal with the domain shift, our method still surpasses all of its counterparts, making a large margin of 1.18 and 1.87 of Dice at 2\% and 5\% labeled data proportion settings. 
Furthermore, when compared with other SOTA pure semi-supervised segmentation methods, \textit{e.g.}, BCP~\cite{bai2023bcp}, our method shows solid performance gains.

\begin{figure*}[ht]
\begin{minipage}[t]{0.45\textwidth}
\makeatletter\def\@captype{table}\makeatother
    \centering
    \caption{Qualitative comparison between \methodName{} and the SOTA methods on 2\% labeled M\&Ms dataset for Semi-MDG task.} \label{mnms}
    \includegraphics[width=0.95\linewidth]{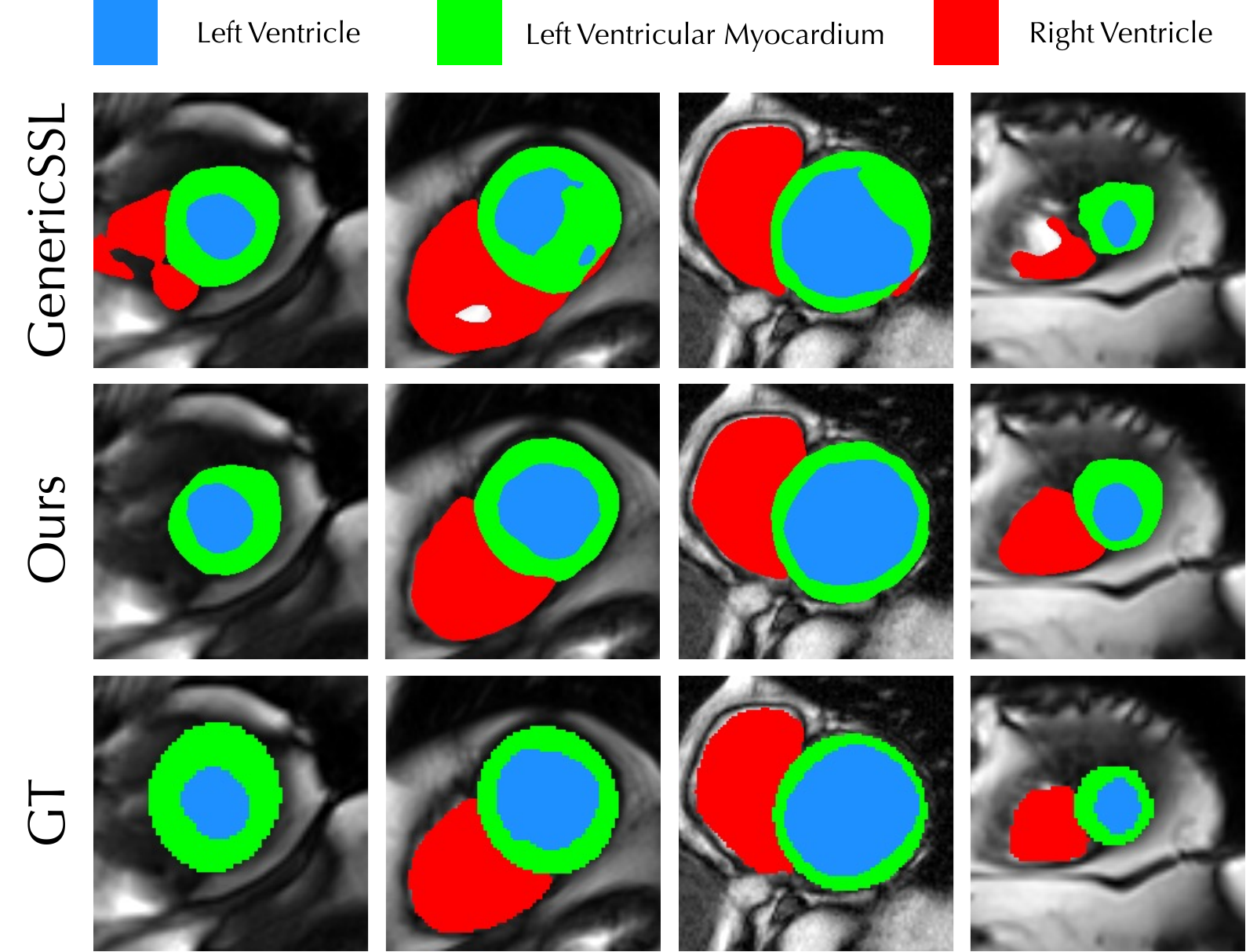}
\end{minipage}\quad
\begin{minipage}[t]{0.54\textwidth}
\makeatletter\def\@captype{table}\makeatother
    \centering
    \caption{T-SNE visualizations of image features of \textcolor{blue}{source domain} and \textcolor{orange}{target domain} produced by our \methodName{} and EPL~\cite{yao2022enhancing} on M\&Ms dataset (test on domain A).} \label{tsne}
    \includegraphics[width=\linewidth]{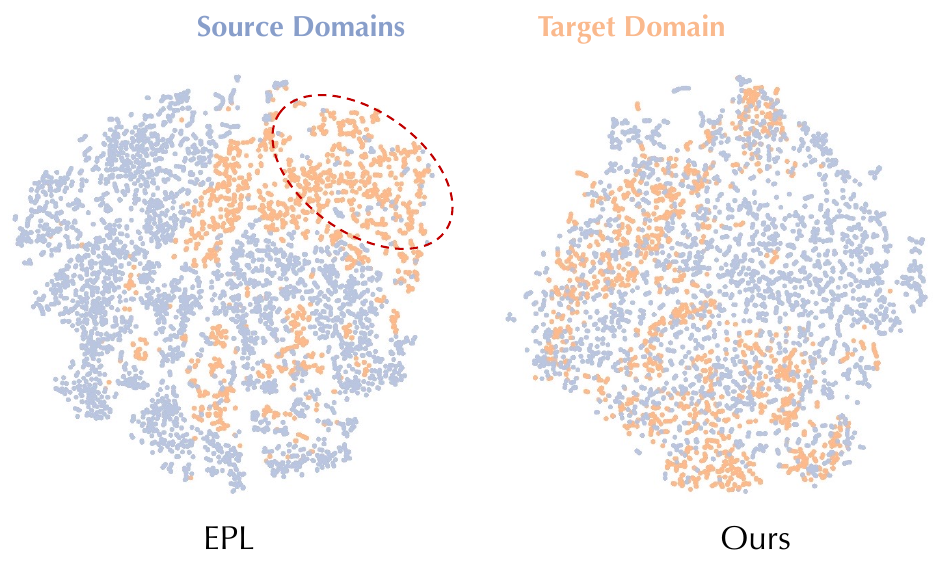}
\end{minipage}\quad
\vspace{-0.6cm}
\end{figure*}
In addition to the quantitative results, we also evaluate our methods qualitatively. Firstly, we conduct a qualitative comparison with GenericSSL~\cite{wang2024towards} which also tried to develop a generic semi-supervised method. As illustrated in Fig.~\ref{mnms}, our method produces more accurate and smooth results which are more similar to the ground truth. Secondly, we compare the T-SNE visualization of the image features produced by our method and SOTA Semi-MDG method, \textit{i.e.}, EPL~\cite{yao2022enhancing}. As we can see in Fig.~\ref{tsne}, compared with EPL, our method successfully reduces the domain distance between the source domains and the unseen target domain.


\begin{table}[!ht]
\caption{Ablations on different segmentation backbones on different scenarios. Many counterparts~\cite{bai2023bcp,yao2022epl,zhou2023semi,liu2021semi,wang2024towards,wang2023dhc,simis,lin2022cld}, suffer performance drops when using Transformer-based backbones, even enlarging the performance gap with our methods in different tasks.}
    \label{ablation:backbone}
    \setlength\tabcolsep{3pt}
    \centering
    \resizebox*{0.48\textwidth}{!}{
    \begin{tabular}{@{}lccccc}
    \toprule
    \rowcolor{red!10}\multicolumn{6}{c}{2\% M\&Ms Dataset (test on domain A) for Semi-MDG task} \\ \midrule
     Backbone & BCP~\cite{bai2023bcp} & EPL~\cite{yao2022epl} & StyleMatch~\cite{zhou2023semi} & Meta~\cite{liu2021semi} & \cellcolor{cyan!20}{\textbf{Ours}} \\ \midrule
     CNN & 71.57& 82.35 & 74.51 & 66.01 & \cellcolor{cyan!20}{N/A} \\
     Transformer & 64.33\textcolor{red}{$\downarrow$} & 79.95\textcolor{red}{$\downarrow$} & 72.33\textcolor{red}{$\downarrow$} & 62.77\textcolor{red}{$\downarrow$} & \cellcolor{cyan!20}{\textbf{86.50}}\\
    \bottomrule
    \bottomrule
    \rowcolor{red!10}\multicolumn{6}{c}{20\% Synapse Dataset for SSMIS task} \\ \midrule
     Backbone & GenericSSL~\cite{wang2024towards} & DHC~\cite{wang2023dhc} & SimiS~\cite{simis} & CLD~\cite{lin2022cld} & \cellcolor{cyan!20}{\textbf{Ours}} \\ \midrule
     CNN & 60.88& 40.07 & 41.07 & 66.01 & \cellcolor{cyan!20}{N/A} \\
     Transformer & 57.39\textcolor{red}{$\downarrow$} & 42.07\textcolor{red}{$\downarrow$} & 34.55\textcolor{red}{$\downarrow$} & 38.63\textcolor{red}{$\downarrow$} & \cellcolor{cyan!20}{\textbf{68.38}}\\ \bottomrule
    \end{tabular}}
    \vspace{-0.7cm}
\end{table}

\begin{table*}[!t]
\caption{Effectiveness of the proposed components on Synapse dataset of the SSMIS task (left) and M\&Ms dataset of the Semi-MDG task (right). We use a basic pseudo-labeling framework~\cite{lee2013pseudo} with the SegFormer~\cite{xie2021segformer} as the baseline, \methodName{} consists of 2 types of knowledge delivery mechanisms, \textit{i.e.}, L2U delivery ($L\to U$) and U2L delivery ($U\to L$).}
\begin{minipage}[t]{0.48\textwidth}
\makeatletter\def\@captype{table}\makeatother
    \centering
    \resizebox*{0.87\linewidth}{!}{
    \begin{tabular}{c|c|c|cc}
    \toprule
\rowcolor{red!10}\multicolumn{5}{c}{20\% labeled Synapse Dataset} \\ \midrule
      Settings & $L\to U$ & $U\to L$ \scriptsize
      & 20\% & 40\% \\\midrule
     Baseline & $\times$  & $\times$ & 25.51 & 25.81 \\
    L2U Only& \checkmark & $\times$ & 37.69  & 40.55 \\
    U2L Only& $\times$ & \checkmark & 61.15  & 64.93 \\
     \rowcolor{cyan!20} \textbf{Ours}  & \checkmark & \checkmark & \textbf{64.46} & \textbf{68.07}\\ 
    \bottomrule
    \end{tabular}}
\end{minipage}\quad
\begin{minipage}[t]{0.48\textwidth}
\makeatletter\def\@captype{table}\makeatother
    \centering
    \resizebox*{0.87\textwidth}{!}{
    \begin{tabular}{c|c|c|cc}
    \toprule
    \rowcolor{red!10}\multicolumn{5}{c}{M\&Ms Dataset (test on Domain A)}  \\ \midrule
     Settings  & $L\to U$ & $U\to L$ & 
        2\% & 5\% \\ \midrule
     Baseline & $\times$  & $\times$ & 58.11 & 72.40 \\
     L2U only & \checkmark & $\times$ & 59.47  & 72.54 \\
     U2L only & $\times$  & \checkmark & 85.02  & 85.52 \\
     \rowcolor{cyan!20} \textbf{Ours}  & \checkmark & \checkmark & \textbf{86.50} & \textbf{87.04}\\ 
    \bottomrule
    \end{tabular}}
\end{minipage}\quad
\label{ablation:components}
\vspace{-0.5cm}
\end{table*}

\subsection{Ablation Study}
\noindent\textbf{Effectiveness of the Knowledge Delivery Mechanism.}
We analyze the effectiveness of different components in \methodName{}, \textit{i.e.}, the Labeled-to-unlabeled (L2U) delivery, and the Unlabeled-to-labeled (U2L) delivery as in Tab.~\ref{ablation:components}. According to the results, the semi-supervised baseline performs badly on both the SSMIS task (Synapse dataset) and the Semi-MDG task (M\&Ms dataset). With the involvement of our proposed L2U knowledge delivery (second row) and U2L knowledge delivery (third row), both the segmentation performance and model generalizability are significantly improved.

\begin{figure}[!t]
\makeatletter\def\@captype{table}\makeatother
    \centering
    \caption{Ablations on hyper-parameters, $\alpha$ (Eq.~\ref{eq:u2l_2}) and $s$ (Eq.~\ref{eq:loss_l2u_1}) on Synapse dataset.}
    \label{ablation:hyparameter}
    \includegraphics[width=0.45\textwidth]{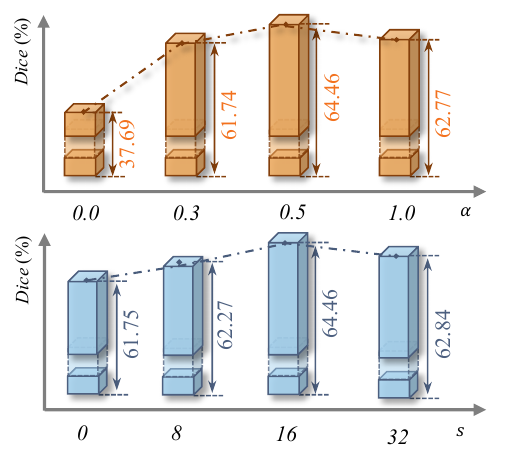}
    \vspace{-0.7cm}
\end{figure}

\noindent\textbf{Ablations on Segmentation Backbones.}
It might be argued that the significant improvements in our method are attributed to the power of the Transformer. Therefore, to address this concern, we further conduct comparison results between our methods and previous Semi-MDG SOTA with the same segmentation backbone. As shown in Fig.~\ref{ablation:backbone}, EPL~\cite{yao2022enhancing} ends up in a performance drop instead of a rise. This is because Transformer requires more sufficient exposure to domain knowledge to enhance its generalizability, due to its limited ability to capture inductive bias~\cite{dosovitskiy2020image}. In contrast, our method facilitates the knowledge delivery across labeled set and unlabeled set which delivers the rich domain knowledge to the learning flow of labeled set, and achieves stable results.

\noindent\textbf{Ablations on Hyper-parameters.}
We also evaluate the performance of our method under different hyper-parameter, \textit{e.g.}, $\alpha$ in Eq.~\ref{eq:u2l_2} and $s$ in Eq.~\ref{eq:loss_l2u_1}, which are shown in Tab.~\ref{ablation:hyparameter} (upper) and Tab.~\ref{ablation:hyparameter} (bottom). As we can observe, our method exhibits minor fluctuations in performance, except for one notable exception when $\alpha$ and $s$ are set to 0, which corresponds to the removal of U2L delivery and L2U delivery respectively.

\section{Conclusion}
This work aims to unify three semi-supervised-related medical image segmentation (SSMIS) tasks, including typical SSMIS, Unsupervised Medical Domain Adaptation, and Semi-supervised Medical Domain Generalization, with a generic SSMIS framework. We identify a critical shared challenge: \textit{the explicit semantic knowledge and rich domain knowledge exclusively exist in the labeled set and unlabeled set respectively}, and propose a \methodName{} to facilitate knowledge delivery across the entire training set. The extensive results on \textbf{six} datasets have presented that with our method, a naive pseudo-labeling scheme can indeed outperform all of the task-tailored SOTAs.

Furthermore, beyond the field of SSMIS, our method also has the potential to benefit the label-efficient training of any Transformer-based framework, including Segment Anything Model~\cite{liu2020saml}. Such broader application potential also sheds light on a wider range of areas.

\bibliographystyle{IEEEtran}
\small{\bibliography{tmi}}

\begin{thebibliography}{10}
\providecommand{\url}[1]{#1}
\csname url@samestyle\endcsname
\providecommand{\newblock}{\relax}
\providecommand{\bibinfo}[2]{#2}
\providecommand{\BIBentrySTDinterwordspacing}{\spaceskip=0pt\relax}
\providecommand{\BIBentryALTinterwordstretchfactor}{4}
\providecommand{\BIBentryALTinterwordspacing}{\spaceskip=\fontdimen2\font plus
\BIBentryALTinterwordstretchfactor\fontdimen3\font minus \fontdimen4\font\relax}
\providecommand{\BIBforeignlanguage}[2]{{%
\expandafter\ifx\csname l@#1\endcsname\relax
\typeout{** WARNING: IEEEtran.bst: No hyphenation pattern has been}%
\typeout{** loaded for the language `#1'. Using the pattern for}%
\typeout{** the default language instead.}%
\else
\language=\csname l@#1\endcsname
\fi
#2}}
\providecommand{\BIBdecl}{\relax}
\BIBdecl

\bibitem{bai2023bcp}
Y.~Bai, D.~Chen, Q.~Li, W.~Shen, and Y.~Wang, ``Bidirectional copy-paste for semi-supervised medical image segmentation,'' in \emph{CVPR}, 2023, pp. 11\,514--11\,524.

\bibitem{gao2023correlation}
S.~Gao, Z.~Zhang, J.~Ma, Z.~Li, and S.~Zhang, ``Correlation-aware mutual learning for semi-supervised medical image segmentation,'' in \emph{International Conference on Medical Image Computing and Computer-Assisted Intervention}.\hskip 1em plus 0.5em minus 0.4em\relax Springer, 2023, pp. 98--108.

\bibitem{wang2023dhc}
H.~Wang and X.~Li, ``Dhc: Dual-debiased heterogeneous co-training framework for class-imbalanced semi-supervised medical image segmentation,'' in \emph{International Conference on Medical Image Computing and Computer-Assisted Intervention}.\hskip 1em plus 0.5em minus 0.4em\relax Springer, 2023, pp. 582--591.

\bibitem{MT}
A.~Tarvainen and H.~Valpola, ``Mean teachers are better role models: Weight-averaged consistency targets improve semi-supervised deep learning results,'' \emph{NIPS}, vol.~30, 2017.

\bibitem{pseudo_label}
D.-H. Lee \emph{et~al.}, ``Pseudo-label: The simple and efficient semi-supervised learning method for deep neural networks,'' in \emph{ICML, Workshops}, vol.~3, 2013, p. 896.

\bibitem{zhu2017cyclegan_uda}
J.-Y. Zhu, T.~Park, P.~Isola, and A.~A. Efros, ``Unpaired image-to-image translation using cycle-consistent adversarial networks,'' in \emph{Proceedings of the IEEE international conference on computer vision}, 2017, pp. 2223--2232.

\bibitem{hoffman2018cycada_uda}
J.~Hoffman, E.~Tzeng, T.~Park, J.-Y. Zhu, P.~Isola, K.~Saenko, A.~Efros, and T.~Darrell, ``Cycada: Cycle-consistent adversarial domain adaptation,'' in \emph{International conference on machine learning}.\hskip 1em plus 0.5em minus 0.4em\relax Pmlr, 2018, pp. 1989--1998.

\bibitem{tsai2018adaouput_uda}
Y.-H. Tsai, W.-C. Hung, S.~Schulter, K.~Sohn, M.-H. Yang, and M.~Chandraker, ``Learning to adapt structured output space for semantic segmentation,'' in \emph{CVPR}, 2018, pp. 7472--7481.

\bibitem{liu2021dgnet}
X.~Liu, S.~Thermos, A.~O’Neil, and S.~A. Tsaftaris, ``Semi-supervised meta-learning with disentanglement for domain-generalised medical image segmentation,'' in \emph{MICCAI}.\hskip 1em plus 0.5em minus 0.4em\relax Springer, 2021, pp. 307--317.

\bibitem{liu2021SDNet}
X.~Liu, S.~Thermos, A.~Chartsias, A.~O’Neil, and S.~A. Tsaftaris, ``Disentangled representations for domain-generalized cardiac segmentation,'' in \emph{Statistical Atlases and Computational Models of the Heart. M\&Ms and EMIDEC Challenges: 11th International Workshop, STACOM 2020, Held in Conjunction with MICCAI 2020, Lima, Peru, October 4, 2020, Revised Selected Papers 11}.\hskip 1em plus 0.5em minus 0.4em\relax Springer, 2021, pp. 187--195.

\bibitem{yao2022epl}
H.~Yao, X.~Hu, and X.~Li, ``Enhancing pseudo label quality for semi-supervised domain-generalized medical image segmentation,'' in \emph{Proceedings of the AAAI Conference on Artificial Intelligence}, vol.~36, 2022, pp. 3099--3107.

\bibitem{liu2022vmfnet}
X.~Liu, S.~Thermos, P.~Sanchez, A.~Q. O’Neil, and S.~A. Tsaftaris, ``vmfnet: Compositionality meets domain-generalised segmentation,'' in \emph{MICCAI}.\hskip 1em plus 0.5em minus 0.4em\relax Springer, 2022, pp. 704--714.

\bibitem{wang2024towards}
H.~Wang and X.~Li, ``Towards generic semi-supervised framework for volumetric medical image segmentation,'' \emph{Advances in Neural Information Processing Systems}, vol.~36, 2024.

\bibitem{yu2019uamt}
L.~Yu, S.~Wang, X.~Li, C.-W. Fu, and P.-A. Heng, ``Uncertainty-aware self-ensembling model for semi-supervised 3d left atrium segmentation,'' in \emph{MICCAI}, 2019, pp. 605--613.

\bibitem{lin2022cld}
Y.~Lin, H.~Yao, Z.~Li, G.~Zheng, and X.~Li, ``Calibrating label distribution for class-imbalanced barely-supervised knee segmentation,'' in \emph{MICCAI}, 2022, pp. 109--118.

\bibitem{lee2013pseudo}
D.-H. Lee \emph{et~al.}, ``Pseudo-label: The simple and efficient semi-supervised learning method for deep neural networks,'' in \emph{Workshop on challenges in representation learning, ICML}, vol.~3.\hskip 1em plus 0.5em minus 0.4em\relax Atlanta, 2013, p. 896.

\bibitem{xie2021segformer}
E.~Xie, W.~Wang, Z.~Yu, A.~Anandkumar, J.~M. Alvarez, and P.~Luo, ``Segformer: Simple and efficient design for semantic segmentation with transformers,'' \emph{Advances in Neural Information Processing Systems}, vol.~34, pp. 12\,077--12\,090, 2021.

\bibitem{zhang2022delving}
C.~Zhang, M.~Zhang, S.~Zhang, D.~Jin, Q.~Zhou, Z.~Cai, H.~Zhao, X.~Liu, and Z.~Liu, ``Delving deep into the generalization of vision transformers under distribution shifts,'' in \emph{CVPR}, 2022, pp. 7277--7286.

\bibitem{chen2021cps}
X.~Chen, Y.~Yuan, G.~Zeng, and J.~Wang, ``Semi-supervised semantic segmentation with cross pseudo supervision,'' in \emph{CVPR}, 2021, pp. 2613--2622.

\bibitem{dosovitskiy2020image}
A.~Dosovitskiy, L.~Beyer, A.~Kolesnikov, D.~Weissenborn, X.~Zhai, T.~Unterthiner, M.~Dehghani, M.~Minderer, G.~Heigold, S.~Gelly \emph{et~al.}, ``An image is worth 16x16 words: Transformers for image recognition at scale,'' \emph{arXiv preprint arXiv:2010.11929}, 2020.

\bibitem{liu2020saml}
Q.~Liu, Q.~Dou, and P.-A. Heng, ``Shape-aware meta-learning for generalizing prostate mri segmentation to unseen domains,'' in \emph{MICCAI}.\hskip 1em plus 0.5em minus 0.4em\relax Springer, 2020, pp. 475--485.

\bibitem{wang2022u2pl}
Y.~Wang, H.~Wang, Y.~Shen, J.~Fei, W.~Li, G.~Jin, L.~Wu, R.~Zhao, and X.~Le, ``Semi-supervised semantic segmentation using unreliable pseudo-labels,'' in \emph{CVPR}, 2022, pp. 4248--4257.

\bibitem{luo2021urpc}
X.~Luo, W.~Liao, J.~Chen, T.~Song, Y.~Chen, S.~Zhang, N.~Chen, G.~Wang, and S.~Zhang, ``Efficient semi-supervised gross target volume of nasopharyngeal carcinoma segmentation via uncertainty rectified pyramid consistency,'' in \emph{MICCAI}, 2021, pp. 318--329.

\bibitem{wu2022cdcl}
H.~Wu, Z.~Wang, Y.~Song, L.~Yang, and J.~Qin, ``Cross-patch dense contrastive learning for semi-supervised segmentation of cellular nuclei in histopathologic images,'' in \emph{CVPR}, 2022, pp. 11\,666--11\,675.

\bibitem{wang2022gbdl}
J.~Wang and T.~Lukasiewicz, ``Rethinking bayesian deep learning methods for semi-supervised volumetric medical image segmentation,'' in \emph{CVPR}, 2022, pp. 182--190.

\bibitem{wu2022ssnet}
Y.~Wu, Z.~Wu, Q.~Wu, Z.~Ge, and J.~Cai, ``Exploring smoothness and class-separation for semi-supervised medical image segmentation,'' in \emph{MICCAI}, 2022, pp. 34--43.

\bibitem{chaitanya2020glcl}
K.~Chaitanya, E.~Erdil, N.~Karani, and E.~Konukoglu, ``Contrastive learning of global and local features for medical image segmentation with limited annotations,'' \emph{Advances in Neural Information Processing Systems}, vol.~33, pp. 12\,546--12\,558, 2020.

\bibitem{dosovitskiy2021vit}
A.~Dosovitskiy, L.~Beyer, A.~Kolesnikov, D.~Weissenborn, X.~Zhai, T.~Unterthiner, M.~Dehghani, M.~Minderer, G.~Heigold, S.~Gelly \emph{et~al.}, ``An image is worth 16x16 words: Transformers for image recognition at scale,'' \emph{ICLR}, 2021.

\bibitem{vitn_cai2023efficientvit}
H.~Cai, C.~Gan, and S.~Han, ``Efficientvit: Enhanced linear attention for high-resolution low-computation visual recognition,'' in \emph{ICCV}, 2023.

\bibitem{cai2022semivit}
Z.~Cai, A.~Ravichandran, P.~Favaro, M.~Wang, D.~Modolo, R.~Bhotika, Z.~Tu, and S.~Soatto, ``Semi-supervised vision transformers at scale,'' \emph{NIPS}, vol.~35, pp. 25\,697--25\,710, 2022.

\bibitem{SemiCVT}
H.~Huang, S.~Xie, L.~Lin, R.~Tong, Y.-W. Chen, Y.~Li, H.~Wang, Y.~Huang, and Y.~Zheng, ``Semicvt: Semi-supervised convolutional vision transformer for semantic segmentation,'' in \emph{CVPR}, 2023, pp. 11\,340--11\,349.

\bibitem{DiverseCotraining}
Y.~Li, X.~Wang, L.~Yang, L.~Feng, W.~Zhang, and Y.~Gao, ``Diverse cotraining makes strong semi-supervised segmentor,'' in \emph{ICCV}, 2023.

\bibitem{huang2023semi}
H.~Huang, Y.~Huang, S.~Xie, L.~Lin, T.~Ruofeng, Y.-w. Chen, Y.~Li, and Y.~Zheng, ``Semi-supervised convolutional vision transformer with bi-level uncertainty estimation for medical image segmentation,'' in \emph{Proceedings of the 31st ACM International Conference on Multimedia}, 2023, pp. 5214--5222.

\bibitem{allspark}
H.~Wang, Q.~Zhang, Y.~Li, and X.~Li, ``Allspark: Reborn labeled features from unlabeled in transformer for semi-supervised semantic segmentation,'' \emph{arXiv preprint arXiv:2403.01818}, 2024.

\bibitem{liu2022act_uda_new}
X.~Liu, F.~Xing, N.~Shusharina, R.~Lim, C.-C. Jay~Kuo, G.~El~Fakhri, and J.~Woo, ``Act: Semi-supervised domain-adaptive medical image segmentation with asymmetric co-training,'' in \emph{MICCAI}.\hskip 1em plus 0.5em minus 0.4em\relax Springer, 2022, pp. 66--76.

\bibitem{dou2019pnp_uda}
Q.~Dou, C.~Ouyang, C.~Chen, H.~Chen, B.~Glocker, X.~Zhuang, and P.-A. Heng, ``Pnp-adanet: Plug-and-play adversarial domain adaptation network at unpaired cross-modality cardiac segmentation,'' \emph{IEEE Access}, vol.~7, pp. 99\,065--99\,076, 2019.

\bibitem{chen2020sifa_uda}
C.~Chen, Q.~Dou, H.~Chen, J.~Qin, and P.~A. Heng, ``Unsupervised bidirectional cross-modality adaptation via deeply synergistic image and feature alignment for medical image segmentation,'' \emph{IEEE transactions on medical imaging}, vol.~39, no.~7, pp. 2494--2505, 2020.

\bibitem{zou2020dsfn_uda}
D.~Zou, Q.~Zhu, and P.~Yan, ``Unsupervised domain adaptation with dual-scheme fusion network for medical image segmentation.'' in \emph{IJCAI}, 2020, pp. 3291--3298.

\bibitem{han2021dsan_uda}
X.~Han, L.~Qi, Q.~Yu, Z.~Zhou, Y.~Zheng, Y.~Shi, and Y.~Gao, ``Deep symmetric adaptation network for cross-modality medical image segmentation,'' \emph{IEEE transactions on medical imaging}, vol.~41, no.~1, pp. 121--132, 2021.

\bibitem{gu2022confuda_uda_new}
M.~Gu, S.~Vesal, M.~Thies, Z.~Pan, F.~Wagner, M.~Rusu, A.~Maier, and R.~Kosti, ``Confuda: Contrastive fewshot unsupervised domain adaptation for medical image segmentation,'' \emph{arXiv preprint arXiv:2206.03888}, 2022.

\bibitem{chen2022dst}
B.~Chen, J.~Jiang, X.~Wang, P.~Wan, J.~Wang, and M.~Long, ``Debiased self-training for semi-supervised learning,'' in \emph{Advances in Neural Information Processing Systems}, 2022.

\bibitem{wang2022depl}
X.~Wang, Z.~Wu, L.~Lian, and S.~X. Yu, ``Debiased learning from naturally imbalanced pseudo-labels,'' in \emph{CVPR}, 2022, pp. 14\,647--14\,657.

\bibitem{guo2022adsh}
L.-Z. Guo and Y.-F. Li, ``Class-imbalanced semi-supervised learning with adaptive thresholding,'' in \emph{ICML}.\hskip 1em plus 0.5em minus 0.4em\relax PMLR, 2022, pp. 8082--8094.

\bibitem{wei2021crest}
C.~Wei, K.~Sohn, C.~Mellina, A.~Yuille, and F.~Yang, ``Crest: A class-rebalancing self-training framework for imbalanced semi-supervised learning,'' in \emph{CVPR}, 2021, pp. 10\,857--10\,866.

\bibitem{simis}
H.~Chen, Y.~Fan, Y.~Wang, J.~Wang, B.~Schiele, X.~Xie, M.~Savvides, and B.~Raj, ``An embarrassingly simple baseline for imbalanced semi-supervised learning,'' \emph{arXiv preprint arXiv:2211.11086}, 2022.

\bibitem{basak2022addressing}
H.~Basak, S.~Ghosal, and R.~Sarkar, ``Addressing class imbalance in semi-supervised image segmentation: A study on cardiac mri,'' in \emph{MICCAI}, 2022, pp. 224--233.

\bibitem{wang2022uctransnet}
H.~Wang, P.~Cao, J.~Wang, and O.~R. Zaiane, ``Uctransnet: rethinking the skip connections in u-net from a channel-wise perspective with transformer,'' in \emph{Proceedings of the AAAI conference on artificial intelligence}, vol.~36, 2022, pp. 2441--2449.

\bibitem{IN}
D.~Ulyanov, A.~Vedaldi, and V.~Lempitsky, ``Instance normalization: The missing ingredient for fast stylization,'' \emph{arXiv preprint arXiv:1607.08022}, 2016.

\bibitem{synapse}
B.~Landman, Z.~Xu, J.~Igelsias, M.~Styner, T.~Langerak, and A.~Klein, ``2015 miccai multi-atlas labeling beyond the cranial vault--workshop and challenge,'' 2015.

\bibitem{xiong2021laseg}
Z.~Xiong, Q.~Xia, Z.~Hu, N.~Huang, C.~Bian, Y.~Zheng, S.~Vesal, N.~Ravikumar, A.~Maier, X.~Yang \emph{et~al.}, ``A global benchmark of algorithms for segmenting the left atrium from late gadolinium-enhanced cardiac magnetic resonance imaging,'' \emph{Medical image analysis}, vol.~67, p. 101832, 2021.

\bibitem{ji2022amos}
Y.~Ji, H.~Bai, C.~Ge, J.~Yang, Y.~Zhu, R.~Zhang, Z.~Li, L.~Zhanng, W.~Ma, X.~Wan \emph{et~al.}, ``Amos: A large-scale abdominal multi-organ benchmark for versatile medical image segmentation,'' \emph{Advances in Neural Information Processing Systems}, vol.~35, pp. 36\,722--36\,732, 2022.

\bibitem{zhuang2016mmwhs}
X.~Zhuang and J.~Shen, ``Multi-scale patch and multi-modality atlases for whole heart segmentation of mri,'' \emph{Medical image analysis}, vol.~31, pp. 77--87, 2016.

\bibitem{prados2017spinal}
F.~Prados, J.~Ashburner, C.~Blaiotta, T.~Brosch, J.~Carballido-Gamio, M.~J. Cardoso, B.~N. Conrad, E.~Datta, G.~D{\'a}vid, B.~De~Leener \emph{et~al.}, ``Spinal cord grey matter segmentation challenge,'' \emph{Neuroimage}, vol. 152, pp. 312--329, 2017.

\bibitem{li2020sassnet}
S.~Li, C.~Zhang, and X.~He, ``Shape-aware semi-supervised 3d semantic segmentation for medical images,'' in \emph{MICCAI}.\hskip 1em plus 0.5em minus 0.4em\relax Springer, 2020, pp. 552--561.

\bibitem{luo2021dtc}
X.~Luo, J.~Chen, T.~Song, and G.~Wang, ``Semi-supervised medical image segmentation through dual-task consistency,'' in \emph{Proceedings of the AAAI Conference on Artificial Intelligence}, vol.~35, 2021, pp. 8801--8809.

\bibitem{wu2021mcnet}
Y.~Wu, M.~Xu, Z.~Ge, J.~Cai, and L.~Zhang, ``Semi-supervised left atrium segmentation with mutual consistency training,'' in \emph{MICCAI}.\hskip 1em plus 0.5em minus 0.4em\relax Springer, 2021, pp. 297--306.

\bibitem{jafari2022lmisa_uda}
M.~Jafari, S.~Francis, J.~M. Garibaldi, and X.~Chen, ``Lmisa: A lightweight multi-modality image segmentation network via domain adaptation using gradient magnitude and shape constraint,'' \emph{Medical Image Analysis}, vol.~81, p. 102536, 2022.

\bibitem{isensee2021nnunet}
F.~Isensee, P.~F. Jaeger, S.~A. Kohl, J.~Petersen, and K.~H. Maier-Hein, ``nnu-net: a self-configuring method for deep learning-based biomedical image segmentation,'' \emph{Nature methods}, vol.~18, no.~2, pp. 203--211, 2021.

\bibitem{li2020lddg}
H.~Li, Y.~Wang, R.~Wan, S.~Wang, T.-Q. Li, and A.~Kot, ``Domain generalization for medical imaging classification with linear-dependency regularization,'' \emph{Advances in Neural Information Processing Systems}, vol.~33, pp. 3118--3129, 2020.

\bibitem{liu2021semi}
X.~Liu, S.~Thermos, A.~O’Neil, and S.~A. Tsaftaris, ``Semi-supervised meta-learning with disentanglement for domain-generalised medical image segmentation,'' in \emph{Medical Image Computing and Computer Assisted Intervention--MICCAI 2021: 24th International Conference, Strasbourg, France, September 27--October 1, 2021, Proceedings, Part II 24}.\hskip 1em plus 0.5em minus 0.4em\relax Springer, 2021, pp. 307--317.

\bibitem{zhou2023semi}
K.~Zhou, C.~C. Loy, and Z.~Liu, ``Semi-supervised domain generalization with stochastic stylematch,'' \emph{International Journal of Computer Vision}, pp. 1--11, 2023.

\bibitem{yao2022enhancing}
H.~Yao, X.~Hu, and X.~Li, ``Enhancing pseudo label quality for semi-supervised domain-generalized medical image segmentation,'' in \emph{AAAI}, vol.~36, 2022, pp. 3099--3107.

\bibitem{xie2023is2net}
S.~Xie, Z.~Niu, H.~Huang, H.~Sun, R.~Qin, Y.-W. Chen, and L.~Lin, ``Is2net: Intra-domain semantic and inter-domain style enhancement for semi-supervised medical domain generalization,'' in \emph{Proceedings of the 31st ACM International Conference on Multimedia}, 2023, pp. 8285--8293.

\bibitem{sohn2020fixmatch}
K.~Sohn, D.~Berthelot, N.~Carlini, Z.~Zhang, H.~Zhang, C.~A. Raffel, E.~D. Cubuk, A.~Kurakin, and C.-L. Li, ``Fixmatch: Simplifying semi-supervised learning with consistency and confidence,'' \emph{Advances in neural information processing systems}, vol.~33, pp. 596--608, 2020.

\end{thebibliography}

\end{document}